\documentclass[runningheads]{llncs}
\usepackage{eccv}
\usepackage{eccvabbrv}
\usepackage{graphicx}
\usepackage{siunitx}
\usepackage{booktabs}
\usepackage{amsmath}
\usepackage{amssymb}
\usepackage{pifont}

\usepackage{tabularx}
\usepackage{adjustbox}
\usepackage[accsupp]{axessibility}
\usepackage[symbol]{footmisc}
\usepackage[breaklinks,colorlinks,citecolor=eccvblue]{hyperref}
\usepackage{orcidlink}
\newcommand{\unnumberedsubsection}[1]{%
  \refstepcounter{subsection}%
  \subsection*{\Alph{subsection}. #1}%
}

\begin{document}
\title{AutoSplat: Constrained Gaussian Splatting for Autonomous Driving Scene Reconstruction} 
\titlerunning{AutoSplat}
\author{Mustafa Khan\inst{1,2,*} \and
Hamidreza Fazlali\inst{2,*} \and
Dhruv Sharma\inst{2}\and \\
Tongtong Cao\inst{2}\and
Dongfeng Bai\inst{2}\and
Yuan Ren\inst{2}\and
Bingbing Liu\inst{2}}
\authorrunning{Khan et al.}
\institute{University of Toronto \and
Noah’s Ark Lab, Huawei Technologies \\}
\maketitle

\begin{abstract}
Realistic scene reconstruction and view synthesis are essential for advancing autonomous driving systems by simulating safety-critical scenarios. 3D Gaussian Splatting excels in real-time rendering and static scene reconstructions but struggles with modeling driving scenarios due to complex backgrounds, dynamic objects, and sparse views. We propose AutoSplat, a framework employing Gaussian splatting to achieve highly realistic reconstructions of autonomous driving scenes. By imposing geometric constraints on Gaussians representing the road and sky regions, our method enables multi-view consistent simulation of challenging scenarios including lane changes. Leveraging 3D templates, we introduce a reflected Gaussian consistency constraint to supervise both the visible and unseen side of foreground objects. Moreover, to model the dynamic appearance of foreground objects, we estimate residual spherical harmonics for each foreground Gaussian. Extensive experiments on Pandaset \cite{pandaset} and KITTI \cite{kitti} demonstrate that AutoSplat outperforms state-of-the-art methods in scene reconstruction and novel view synthesis across diverse driving scenarios. Our project page is at: \url{https://autosplat.github.io/}
\end{abstract}

\section{Introduction}
\label{sec:intro}
\footnote[0]{*Denotes equal contribution.}
View synthesis and scene reconstruction from captured images are fundamental challenges in computer graphics and computer vision \cite{hartley, colmap, colmap2}, crucial for autonomous driving and robotics. Reconstructing detailed 3D scenes from sparse sensor data on moving vehicles \cite{pandaset, kitti} is especially challenging at high speeds, where both the ego-vehicle and surrounding objects are in motion. These techniques enhance safety by simulating realistic driving scenarios, particularly for costly or hazardous corner cases. 

The advent of Neural Radiance Fields (NeRFs) \cite{nerf}  transformed view synthesis and reconstruction by implicitly representing a scene using a multi-layer perceptron (MLP).  Numerous efforts have addressed NeRF's challenges, such as slow training and rendering speed \cite{speed, ngp, tensorf, factor, mobilenerf, merf, bakedsdf}, as well as rendering quality \cite{mipnerf, refnerf, zipnerf}, particularly in reconstructing bounded static scenes. Extensions to unbounded scenes and large-scale urban areas have also been explored \cite{wildnerf, mipnerf360, blocknerf, meganerf, streetsurf}. Various methods have addressed dynamic scene modeling in autonomous driving scenarios \cite{nsg, nsg2, mars, unisim}. Yet, NeRF-based methods still face significant hurdles in training and rendering large-scale scenes with multiple dynamic objects.

\begin{figure}[t]
  \centering
  \includegraphics[height=4.5cm]{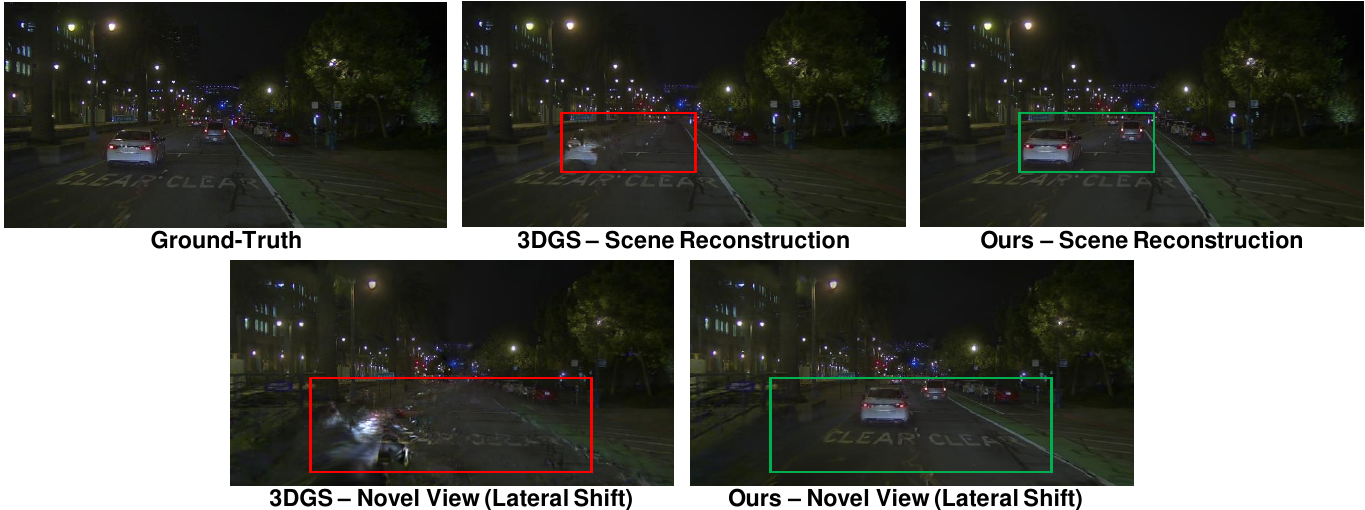}
  \caption{Scene reconstruction and novel view synthesis (ego-vehicle lane change) results of 3DGS \cite{3dgs} and our proposed method on a sample autonomous driving scene.}
  \label{fig:3dgsproblems}
\end{figure}

In contrast to NeRF-based methods, 3D Gaussian Splatting (3DGS) \cite{3dgs} explicitly represents the scene using anisotropic 3D Gaussians, which enables faster training, achieves high-quality novel view synthesis, and real-time rasterization. Despite its proficiency in handling purely static scenes, 3DGS is unable to reconstruct scenes with dynamic objects. Moreover, 3DGS is not designed for reconstructing autonomous driving scenes, where sparse views are available. These result in distortions in foreground object reconstruction and novel view synthesis, such as the ego-vehicle lane change scenario shown in \cref{fig:3dgsproblems}.

In this paper, we propose AutoSplat, a specially designed 3DGS-based framework for the simulation of autonomous driving scenes. To ensure consistent and high-quality synthesis in novel views during background reconstruction, we distinguish between the road and sky regions from the rest of the background. We constrain their Gaussians to become flat, guaranteeing multi-view consistency. This is especially evident in lane change scenarios, as illustrated in \cref{fig:3dgsproblems}. Additionally, the 3D points representing foreground objects are not captured by structure-from-motion (SfM) methods and the LiDAR point clouds are sparse and incomplete. Therefore, we leverage a dense 3D template as a prior for initialization of Gaussians, which are fine-tuned to reconstruct the foreground objects in the scene. This allows us to introduce the reflected Gaussian consistency constraint, which supervises the unseen  portion of a foreground object by reflecting all Gaussians across its symmetric plane using the ground-truth camera views. Finally, to capture the dynamic appearance of foreground objects, we estimate residual spherical harmonics per Gaussian across different time steps. Overall, our key contributions are four-fold:
\begin{itemize}
    \item Decomposing the background and geometrically constraining its road and sky regions to enable multi-view consistent rasterization.
    \item Leveraging 3D templates for initializing foreground Gaussians paired with a reflected Gaussian consistency constraint to reconstruct unseen parts from symmetrically visible views. 
    \item Capturing the dynamic visual characteristics of foreground objects through the estimation of temporally-dependent, residual spherical harmonics. 
    \item We comprehensively evaluate AutoSplat against state-of-the-art (SOTA) methods on Pandaset \cite{pandaset} and KITTI \cite{kitti}. Furthermore, extensive ablation studies demonstrate the effectiveness of our proposed components.
\end{itemize}

\section{Related Work}
\label{sec:relwork}

\subsubsection{Implicit Representations and Neural Rendering} 
Volumetric rendering techniques, notably NeRF, have significantly advanced 3D reconstruction and novel view synthesis. However, NeRF encounters challenges including slow training and rendering, high memory usage, and imprecise geometry estimation, particularly with sparse viewpoints \cite{volumetric, nerf, ngp}.
To address the slow training speed different approaches such as voxel grids \cite{voxel, plenoxel}, tensor factorization \cite{tensorf, factor} as well as hash encoding \cite{ngp, suds}, have been explored. For improving the rendering latency, FasterNeRF \cite{fastnerf}, devised a graphic-inspired factorization to compactly cache a deep radiance map at each position in space, while efficiently querying that map using ray directions. 
MobileNeRF \cite{mobilenerf} and BasedSDF \cite{bakedsdf}, achieve fast rendering speed by transforming implicit volumes into explicit textured meshes. To tackle the low-quality rendering of NeRF, Mip-NeRF \cite{mipnerf}, efficiently renders anti-aliased, conical frustums instead of rays. Mip-NeRF 360 \cite{mipnerf360}, addresses the inherent ambiguity of large (unbounded) scenes from a small set of images by employing a non-linear scene parameterization, online distillation, and a distortion-based regularizer.

\subsubsection{Urban Scene Reconstruction with NeRF} Modeling city-scale scenes is challenging due to managing thousands of images with varied lighting conditions, each capturing only a fraction of the scene, posing significant computational demands. MegaNeRF \cite{meganerf} and BlockNeRF \cite{blocknerf} partition the scene into blocks and train separate NeRF models for each block. However, these approaches do not model dynamic objects conventionally found in autonomous driving scenarios. NSG \cite{nsg} and MARS \cite{mars} perform dynamic scene modeling by incorporating a scene graph. Unlike NSG, SUDS \cite{suds} addresses reconstruction during ego-vehicle motion, utilizing LiDAR data for improved depth perception and optical flow to alleviate the stringent demand for object labeling. EmerNeRF \cite{emernerf} learns spatial-temporal representations of driving scenarios by stratifying scenes and using induced flow fields to enhance rendering precision of dynamic objects. Despite optimization efforts and innovative strategies, NeRF-based approaches remain computationally demanding and necessitate densely overlapping views. Moreover, constraints on model capacity pose challenges in accurately modeling long-term dynamic scenes with multiple objects, resulting in visual artifacts.

\subsubsection{3D Gaussian Splatting (3DGS)} 3DGS \cite{3dgs} utilizes an explicit scene representation. Central to its effectiveness is the optimization of anisotropic 3D Gaussians, responsible for the faithful reconstruction of a scene, complemented by the integration of a swift, visibility-aware rasterization algorithm. This not only expedites training but also facilitates real-time rasterization. However, 3DGS still faces considerable hurdles in reconstructing large-scale autonomous driving scenes due to its static-scene assumption and the availability of limited camera views. Moreover, the absence of geometric constraints on background regions in 3DGS leads to a considerable degradation in quality when synthesizing novel views, as shown in \cref{fig:3dgsproblems}. Recently, PVG \cite{pvg}, built on 3DGS to model dynamic scenarios in autonomous driving scenes by using periodic vibration-based temporal dynamics. However, this method does not tackle the simulation of novel scenarios, such as ego-vehicle lane changes and adjusting object trajectories. In contrast, our approach excels in reconstructing dynamic scenes and simulating diverse novel scenarios, including altering the trajectories of both the ego-vehicle and foreground objects.

\section{Method}
\label{sec:method}

\subsection{Prerequisites}
3DGS \cite{3dgs} explicitly represents a scene using anisotropic 3D Gaussians initialized from a set of 3D points. It is defined as:  

\begin{equation}
G(x) = e^{-\frac{1}{2}{(x-\mu)}^T\Sigma^{-1}(x-\mu)}
\label{eq:gaussian}
\end{equation}
where, $\mu \in \mathbb{R}^3$ and  $\Sigma \in \mathbb{R}^{3\times3}$ represent the center vector and covariance matrix of each 3D Gaussian, respectively. Moreover, in 3DGS \cite{3dgs}, each Gaussian is assigned an opacity $o$ and color $c$ attributes, where the latter is represented using spherical harmonic coefficients $f_{SH}$. For ease of optimization, the covariance matrix $\Sigma$ is decomposed into a scaling matrix $S$ and a rotation matrix $R$:

\begin{equation}
\label{eq:cov}
\Sigma = RSS^TR^T
\end{equation}

For differentiable rendering, the 3D Gaussians are splatted onto the image plane by approximating their projected position and covariance in 2D. By sorting the Gaussians according to their depth in camera space, each attribute of the Gaussian is queried and the final rasterized color $C$ of a pixel is computed by blending the contributions of $N$ overlapping Gaussians as:

\begin{equation}
C = \sum_{i=1}^{N} c_i\sigma_i\prod_{j=1}^{N-1}(1-\sigma_j), \quad \sigma_i=o_iG'_i(x')
\end{equation}
where, $x'$ is the target pixel position and $G'_i$ is the $i$th splatted Gaussian. Leveraging the differentiable rasterizer, the five learnable attributes of the 3D Gaussians ($\mu, o, c, S, R$), are directly optimized using training-view reconstruction.

\subsection{Overview}
\begin{figure}[t]
  \centering
  \includegraphics[height=3.9cm]{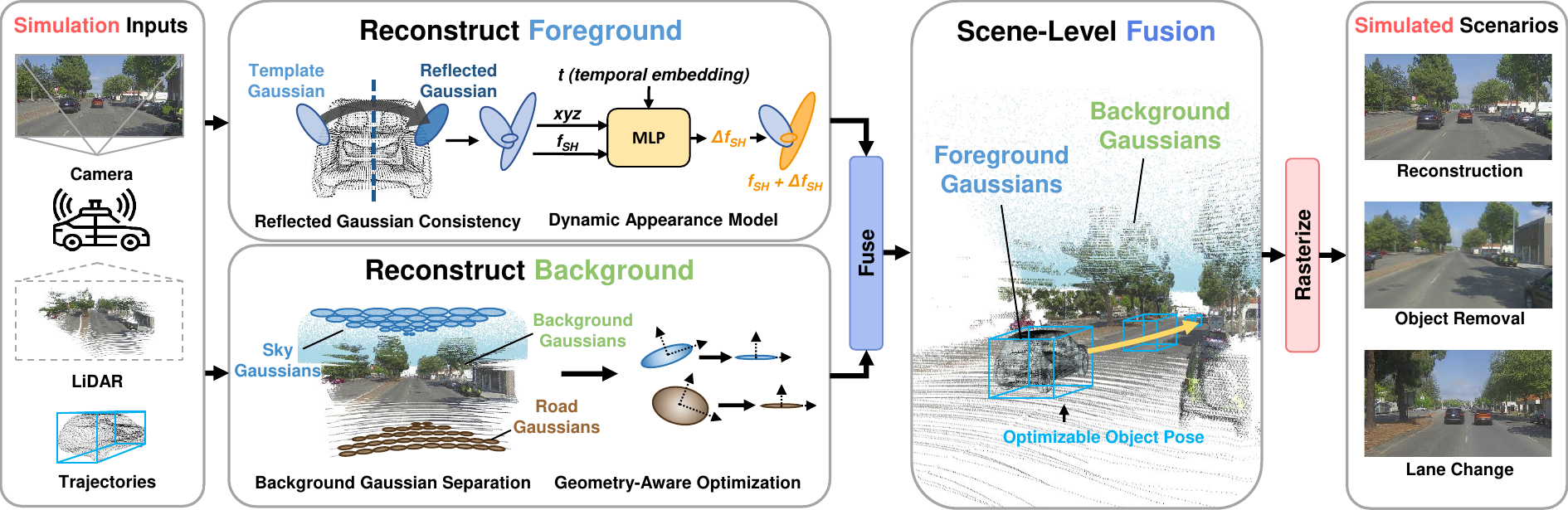}
  \caption{\textbf{AutoSplat overview.} The proposed framework reconstructs the background and foreground separately and then fuses them to simulate different scenarios.}
  \label{fig: architecture}
\end{figure}

Given sequentially captured and calibrated multi-sensor data, which comprises a series of $N$ images ($I_i$) taken by a camera with its corresponding intrinsic ($K_i$) and extrinsic ($E_i$) matrices, along with the 3D LiDAR point clouds $L_i$ and corresponding dynamic objects trajectories $T_i$, our objective is to leverage 3DGS to reconstruct the 3D scene and synthesize novel views at any camera pose with new object trajectories.  The overview of our proposed method is shown in \cref{fig: architecture}. We begin by reconstructing a geometry-aware, static background. Then, from a 3D template, foreground objects are reconstructed, ensuring consistency between visible and unseen regions while modeling their dynamic appearances. Finally, we fuse the foreground and background Gaussians to produce a refined and unified representation.

\subsection{Background Reconstruction}
\label{sec:bg-reconstruction}

Autonomous driving scenes are large and unbounded, while sensor observations are sparse. Naively using 3DGS to represent the background from these limited observations is insufficient for realistic reconstruction and simulation. Furthermore, the Gaussians reconstructing the road and sky regions suffer from being geometrically incorrect and produce floater artifacts. While these Gaussians are capable of reconstructing the scene from ground-truth views, their incorrect geometry produces distortions evident when simulating novel scenarios such as laterally shifting the ego-vehicle as shown in \cref{fig:3dgsproblems}.

To address these issues, the background training in our framework is conducted in two phases. In the first phase, the road and sky regions are decomposed from the rest of the background using semantic masks obtained from an off-the-shelf, pre-trained segmentation model \cite{mask2former}. By projecting LiDAR points to the image plane at each time step $i$ using the calibration matrices, each Gaussian is assigned to one of the road, sky, and other class. The aim of this decomposition is two-fold. First, this prevents non-sky and non-road Gaussians from reconstructing the sky and road regions. Second, the sky and road Gaussians can be constrained to produce multi-view consistent results when splatted. Since LiDAR points do not include sky points, we add a plane of points representing the sky above the maximum scene height. The aforementioned regions are supervised using the $L_{L1}$ and $L_{DSSIM}$ loss terms as in \cite{3dgs}. To ensure consistency across views when splatting road and sky Gaussians, these Gaussians are constrained to be flat. This is obtained by minimizing their roll and pitch angles as well as their vertical scale. Therefore, the overall loss terms of background training in the first phase are defined as:
\begin{equation}
     \mathcal{L}_{BG} = (1-\lambda)\mathcal{L}_{1}(I_{g}, \hat{I}_{g}) + \lambda \mathcal{L}_{DSSIM}(I_{g}, \hat{I}_{g}) + \beta\mathcal{C}_{g} \quad g \in \{road, sky, other\}
\end{equation}
\begin{equation}
  \mathcal{C}_{g}=\begin{cases}
    \frac{1}{N_{g}} \sum_{i=1}^{N_{g}} \left( |\phi_i| + |\theta_i| + |s_{z_i}| \right) & \text{if } g \in \{road, sky\} \\
    0 & \text{else }
  \end{cases}
\end{equation}
where $I_{g}$ and $\hat{I}_{g}$, represent the semantically masked ground-truth and rasterized images for region $g$, which can be road, sky or other. $\mathcal{C}_{g}$ is the constraint applied on the road and sky regions, in which $\phi_i$, $\theta_i$ and $s_{z_i}$ denote the roll and pitch angles as well as the vertical scale (along the Z-axis) of the $i$th Gaussian. Additionally, $\beta$ is used to weight the geometry constraint. The proposed constraint guarantees consistent rasterization of road and sky Gaussians regardless of changes in viewpoint. 

In the second phase of background reconstruction, all Gaussians are splatted together and supervised on the whole image using $\mathcal{L}_{BG}$ with $g \in \{road \cup sky \cup other\}$. 
During this phase, the road, sky, and other regions of the background are blended to optimize the final background image. It needs to be mentioned that, in both phases of training, dynamic foreground regions are masked out.

\subsection{Foreground Reconstruction}
\label{sec:fg-reconstruction}
Foreground reconstruction in autonomous driving scenes is crucial for realistic simulation despite challenges like occlusions and dynamic appearances. Here, we introduce novel strategies to tackle these complexities in the 3DGS paradigm.

\subsubsection{Constructing Template Gaussians}
3DGS faces challenges in reconstructing foreground objects due to its dependence on SfM techniques tailored for static scenes and its lack of motion modeling capabilities. To overcome these limitations, we need an alternative approach to initialize the Gaussians representing these foreground objects and optimize their properties. This can be done by leveraging randomly initialized points, accumulated LiDAR scans, or using single or few-shot 3D reconstruction methods \cite{eg3d, nfi, pointe}. Although LiDAR captures detailed geometry, it has limitations including blind spots and sparse surface details for distant objects. Therefore, we use a 3D template with realistic vehicle geometry to model the foreground objects. Notably, we employ \cite{nfi}, which generates 3D shapes of objects such as vehicles from a single image. In our approach, given a sequence of frames with $K$ foreground objects, the template is copied $K$ times and placed into the scene based on the object trajectories. The Gaussians of each foreground object are initialized from this template and scaling factors along each axis are computed to adjust the size of the template to match the dimensions of the target object's 3D bounding box. During training, the associated Gaussians of these templates are iteratively optimized to converge to the target appearances. By exploiting the rich geometric information encoded in the templates, our method enhances the realism and fidelity of the foreground reconstruction. At the same time, we retain explicit control over the placement of template Gaussians, allowing us to generate new scenarios by modifying the trajectories of foreground objects.

\paragraph{\small \normalfont \textbf{Reflected Gaussian Consistency}} 
\begin{figure}[t]
  \centering
  \includegraphics[height=2.6cm]{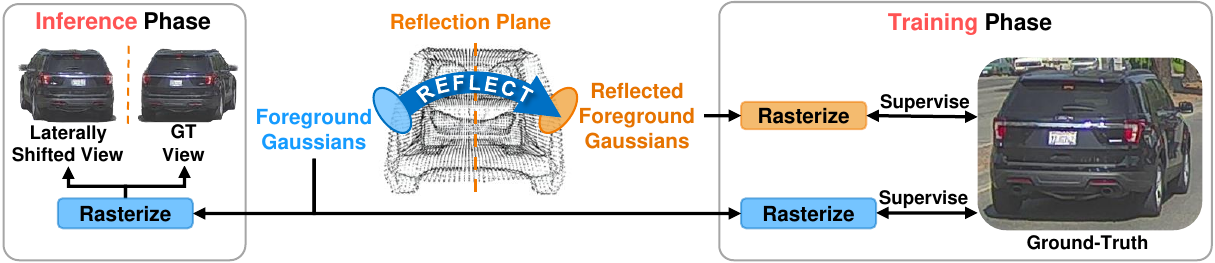}
  \caption{Reflected Gaussian consistency in training and inference phases.}
  \label{fig:reflected_consistency}
\end{figure}
Foreground objects exhibit symmetry in their structure. Leveraging this assumption helps to improve the reconstruction quality, especially in scenarios with limited views \cite{neusim}. We broaden the application of this assumption within the 3DGS paradigm by enforcing consistency between visible and symmetrically unseen sides of the foreground object. This process is shown in \cref{fig:reflected_consistency}. More specifically, for each foreground object, Gaussians are reflected across the object's plane of symmetry. The reflected Gaussians are then rasterized and supervised based on the ground-truth view. This will provide supervision for Gaussians that are not visible. The reflection matrix $M$ for the Gaussians can be defined as:
\begin{equation}
    M = I - 2 \frac{aa^T}{\|a\|^2}
\end{equation}
where $a$ represents the axis of reflection and $I$ is identity. The position $x$, rotation $R$, spherical harmonic features $f_{SH}$ of each Gaussian are reflected by:
\begin{equation}
    \begin{split}
        \tilde{x} &= Mx \\
        \tilde{R} &= MR \\
        \tilde{f}_{SH} &= D_{M}f_{SH}
    \end{split}
\end{equation}
where $D_M$ is a Wigner D-matrix describing a reflection and $\tilde{x}$, $\tilde{R}$, $\tilde{f}_{SH}$ represent the reflected position, rotation, and spherical harmonic features of the Gaussians, respectively. This reflected consistency constraint enforces the rendering results of the Gaussian of the two symmetrical sides of the object to be similar. During the inference phase, this enables our method to rasterize high-quality foregrounds on their symmetrical views.

\paragraph{\small \normalfont \textbf{Dynamic Appearance Modeling}} Capturing the dynamic appearance of foreground objects is vital for autonomous driving simulations. This includes vital signals such as indicator lights, headlights, and taillights, which communicate intentions and influence driving behavior. Additionally, realistic simulation necessitates modeling various changes in lighting conditions such as shadows.

To capture dynamic appearance, we learn a 4D representation of the foreground objects' appearance by learning residual spherical harmonic features for each Gaussian. In other words, the estimated residual features are used to impart dynamic appearance onto static representations. Here, a simple MLP is used to model dynamic appearance. More concretely, we utilize temporal embeddings, recognizing that change in appearance is intricately linked to temporal evolution.
At each time step, the corresponding temporal embeddings, Gaussian positions, and spherical harmonic features are fed to the model. Then, the estimated residual features are added to the original spherical harmonic features. Therefore, the dynamic appearance of the foreground object at each time step is modeled by:
\begin{equation}
    \begin{split}
        \Delta f_{SH, t} &= MLP(E_{t}, x, f_{SH}) \\
        f_{SH, t} &= f_{SH} + \Delta f_{SH, t}
    \end{split}
\end{equation}
where $t$, is the time step, $E_{t}$ is the corresponding temporal embeddings. Also, $x$ and $f_{SH}$ represent the Gaussian positions and spherical harmonic features, respectively. $\Delta f_{SH, t}$ and $f_{SH, t}$ denote the estimated residual and final spherical harmonic features at time step $t$, respectively. This allows the appearance of the foreground objects to be disentangled into a static $f_{SH}$ and dynamic $\Delta f_{SH}$ component. Moreover, incorporating the Gaussian positions as an input to the MLP is crucial for learning position-dependent offsets, which localizes changes in appearance on the foreground object. Furthermore, as elaborated in \cref{sec:exp}, the estimated residual component adds some appearance details. We additionally constrain these offsets to be sparse, preventing flickering artifacts.

The overall loss for optimizing the foreground reconstruction is as follows:
\begin{equation}
    \begin{aligned}
         \mathcal{L}_{FG} = (1-\lambda)\mathcal{L}_{1}(I_{g}, \hat{I}_{g}) + \lambda \mathcal{L}_{DSSIM}(I_{g}, \hat{I}_{g}) + (1-\lambda)\mathcal{L}_{1}(I_{g}, \tilde{I}_{g}) +  \\  \lambda \mathcal{L}_{DSSIM}(I_{g}, \tilde{I}_{g}) + \gamma\mathcal{L}_{1}(\Delta \boldsymbol{f}_{SH, t}) \quad g \in \{fg_{1}, fg_2, ..., fg_{K}\}
     \end{aligned}
\end{equation}
where $K$ is the number of foreground objects, and $fg_{k}$ is a set of Gaussians representing the $k$th object. $I_{g}$, $\hat{I_{g}}$, and $\tilde{I_{g}}$ denote masked ground-truth image, rasterized images of foreground Gaussians and reflected foreground Gaussians, respectively. Also, the sparsity constraint applied to the estimated residual spherical harmonic features is weighted by $\gamma$.

\subsection{Scene-Level Fusion}
Scene-level fusion comprises blending the foreground and background Gaussians. When separately optimized, these two sets of Gaussians exhibit distortions when rasterized together, particularly evident near the foreground objects' borders. 

To address these distortions, both the foreground and background Gaussians are fine-tuned together and supervised on the whole image. This will result in a fused foreground-background image, in which the distortions of both components are alleviated. Moreover, to address the noisy object trajectories, we optimize a transformation correction per object, comprising rotation and translation offsets. These are applied to foreground object tracks to overcome noisy 3D bounding boxes. The final loss term is obtained as:
\begin{equation}
\mathcal{L} = \mathcal{L}_{BG} + \mathcal{L}_{FG}
\end{equation}

\section{Experiments}
\label{sec:exp}
We present the experimental setup in \cref{sec:es}, followed by a comparative evaluation of our approach against SOTA methods using publicly available datasets in \cref{sec:mr}. Additionally, we conduct a comprehensive examination of the proposed strategies to elucidate their effectiveness and potential advantages in \cref{sec:as}.

\subsection{Experimental Setup}
\label{sec:es}

\subsubsection{Dataset} We experimented with two open-source self-driving datasets, KITTI \cite{kitti} and Pandaset \cite{pandaset}. For KITTI, our approach closely followed existing methods. Pandaset includes 103 urban driving scenarios in San Francisco, each with 80 image frames and corresponding LiDAR point clouds. We selected 10 challenging sequences with various dynamic scenes, including multiple foreground objects as well as day and night scenarios.

\subsubsection{Evaluation Metric} When synthesizing novel views with ground-truth images, we use standard metrics such as PSNR, SSIM \cite{ssim}, and LPIPS \cite{lpips} for quantitative evaluations. However, for novel view synthesis with lateral ego-vehicle trajectory adjustments, we report FID \cite{fid}.

\subsubsection{Implementation Details} 
Our implementation is based on the 3DGS framework \cite{3dgs}. Instead of SfM points, we use accumulated LiDAR point clouds for background initialization. Following 3DGS, our background training includes 30K iterations split into two phases of 15K iterations each. Foreground training comprises 5K iterations, followed by 10K iterations for scene fusion, where both foreground and background Gaussians are fine-tuned together. During fusion, attributes of the foreground object Gaussians are fined-tuned while for the background Gaussians adjustments are solely made to the opacity, as their appearance and geometry are established during the background training phase. More details are provided in the supplementary materials.

\subsection{Main Results} 
\label{sec:mr}

\renewcommand{\arraystretch}{1.2}
\begin{table}[tb]
  \caption{Scene reconstruction and novel view synthesis performance comparison on Pandaset.}
  \label{tab:pandaset1}
  \centering
  \resizebox{1\columnwidth}{!}{
  \begin{tabular}{@{} c @{\hspace{1cm}} c @{\hspace{1cm}} c @{\hspace{1cm}} c @{\hspace{1cm}} c @{\hspace{1cm}} c @{\hspace{1cm}} c @{\hspace{1cm}} c @{}}
    \toprule
    & & \multicolumn{3}{c}{\textbf{Scene Reconstruction}} & \multicolumn{3}{c}{\textbf{Novel View Synthesis}} \\
    \cmidrule(r){3-5} \cmidrule(l){6-8}
    \textbf{Methods} & \textbf{FPS} $\uparrow$ & \textbf{PSNR} $\uparrow$ & \textbf{SSIM} $\uparrow$ & \textbf{LPIPS} $\downarrow$ & \textbf{PSNR} $\uparrow$ & \textbf{SSIM} $\uparrow$ & \textbf{LPIPS} $\downarrow$ \\
    \midrule
    NSG \cite{nsg} & 0.021 & 22.99 & 0.678 & 0.569 & 22.79 & 0.802 & 0.578 \\
    MARS \cite{mars} & 0.045 & 23.42 & 0.717 & 0.492 & 23.66 & 0.832 & 0.502 \\
    SUDS \cite{mars} & 0.016 & \textbf{30.19} & 0.910 & 0.355 & 25.13 & 0.843 & 0.426 \\
    EmerNeRF \cite{emernerf} & 0.062 & 29.96 & 0.832 & 0.368 & 27.73 & 0.801 & 0.394 \\
    \midrule
    Ours & \textbf{26} & 29.691 & \textbf{0.936} & \textbf{0.259} & \textbf{27.84} & \textbf{0.906} & \textbf{0.291} \\
    \bottomrule
  \end{tabular}}
\end{table}

\begin{figure}[tb]
  \centering
  \includegraphics[height=4.2cm]{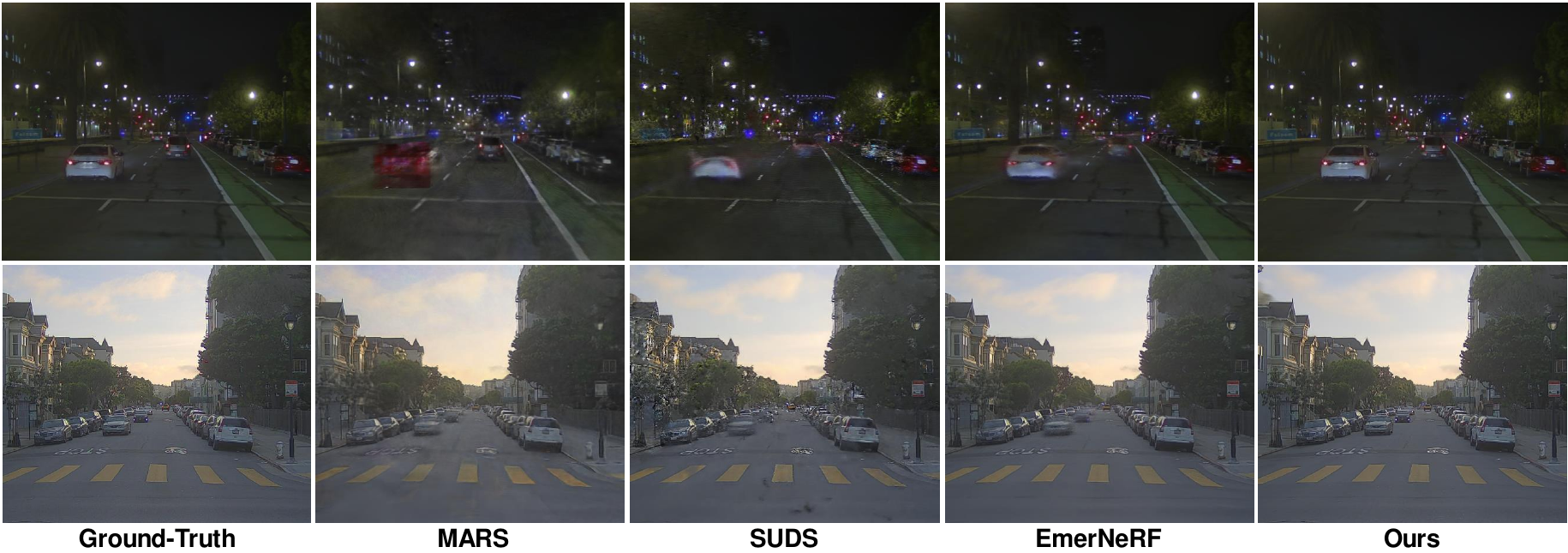}
  \caption{Qualitative comparison of novel view synthesis (test frames) on Pandaset.} 
  \label{fig:pandasetnvs}
\end{figure}

In Pandaset experiments, we benchmark our method against SOTA methods for scene reconstruction and novel view synthesis tasks. Unlike novel view synthesis, where 10\% of frames are excluded, all frames are used during training for scene reconstruction, representing upper-bound results. These results are presented in \cref{tab:pandaset1}. Our method shows significant superiority over alternatives across various evaluation metrics, notably excelling in SSIM and LPIPS. While achieving similar PSNR to EmerNeRF, our method notably outperforms SUDS in novel view synthesis, with slightly reduced scene reconstruction performance. Additionally, our method offers faster execution speed, highlighting its efficiency for real-world applications. This evidence demonstrates the efficacy and robustness of our approach, making it a compelling solution for high-quality reconstructions and realistic synthesis. The qualitative novel view synthesis results on different scenes are shown in \cref{fig:pandasetnvs}. Comparing the results, it can be seen that our method demonstrates exceptional capacity to generate more realistic renderings.

\begin{table}[tb]
  \caption{Quality comparison using Fréchet Inception Distance (FID $\downarrow$) on Pandaset}
  \label{tab:lateralshift}
  \centering
  \scriptsize
  \setlength{\tabcolsep}{4pt}
  \begin{tabular}{@{}lccccc@{}}
    \toprule
    \multicolumn{1}{c}{\textbf{Lateral Shift}} & \textbf{NSG}\cite{nsg} & \textbf{MARS}\cite{mars} & \textbf{SUDS}\cite{suds} & \textbf{EmerNeRF}\cite{emernerf} & \multicolumn{1}{c}{\textbf{Ours}}\\
    \midrule
    \multicolumn{1}{c}{1 Meter}  & 259.9 & 151.8  & 95.4 & 68.2 & \multicolumn{1}{c}{\textbf{54.7}}\\
    \multicolumn{1}{c}{2 Meters}  & 268.7 & 158.9  & 122.7 & 90.4 & \multicolumn{1}{c}{\textbf{68.7}}\\
    \multicolumn{1}{c}{3 Meters}  & 272.8 & 181.9  & 150.8 & 102.8 & \multicolumn{1}{c}{\textbf{83.0}}\\
    \bottomrule
  \end{tabular}
\end{table}

\begin{figure}[tb]
  \centering
  \includegraphics[height=3.9cm]{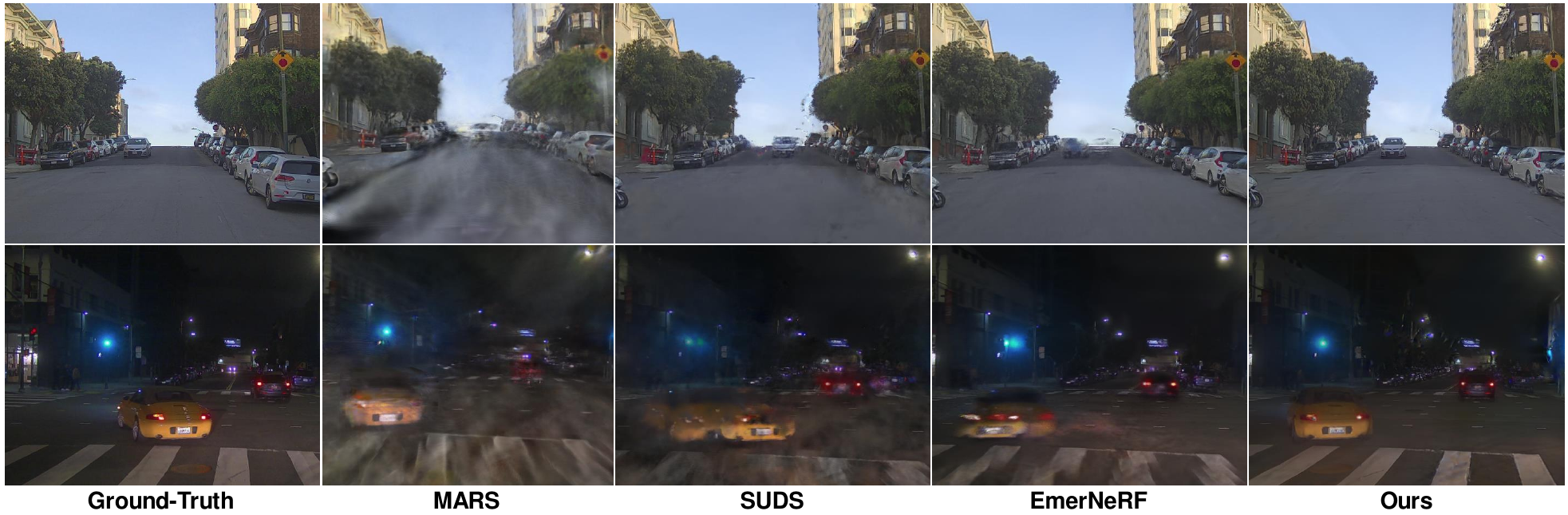}
  \caption{Qualitative comparison of ego-vehicle lateral shift (2 meters) on Pandaset.} 
  \label{fig:pandasetlateralshift}
\end{figure}

Simulating lane changes is a crucial aspect of replicating real-world scenarios in autonomous driving environments. We comprehensively evaluated our approach against competitors by assessing scene realism across varying degrees of lateral shift for the ego-vehicle, using FID. Results in \cref{tab:lateralshift} show that our method consistently outperforms alternatives, demonstrating superior synthesis quality. These findings highlight our approach's efficacy in capturing complex driving maneuvers, attributed to improved background-foreground modelling. Specifically, the geometric and reflected Gaussian consistency constraints enabled multi-view consistent rasterization. Sample qualitative results are shown in \cref{fig:pandasetlateralshift}, which demonstrate the outstanding visual quality of our method.

\begin{table}[tb]
  \caption{Novel view synthesis performance comparison on KITTI.}
  \label{tab:kitti}
  \centering
  \resizebox{\columnwidth}{!}{
  \begin{tabular}{@{} cccccccccc @{}}
    \toprule
    & \multicolumn{3}{c}{\textbf{KITTI - 75\%}} & \multicolumn{3}{c}{\textbf{KITTI - 50\%}} & \multicolumn{3}{c}{\textbf{KITTI - 25\%}} \\
    \cmidrule(lr){2-4} \cmidrule(lr){5-7} \cmidrule(lr){8-10}
    \multicolumn{1}{c}{\textbf{Methods}} & \textbf{PSNR} $\uparrow$ & \textbf{SSIM} $\uparrow$ & \textbf{LPIPS} $\downarrow$ & \textbf{PSNR} $\uparrow$ & \textbf{SSIM} $\uparrow$ & \textbf{LPIPS} $\downarrow$ & \textbf{PSNR} $\uparrow$ & \textbf{SSIM} $\uparrow$ & \textbf{LPIPS} $\downarrow$ \\
    \midrule
    NeRF \cite{nerf} & 18.56 & 0.557 & 0.554 & 19.12 & 0.587 & 0.497 & 18.61 & 0.570 & 0.510 \\
    NSG \cite{nsg} & 21.53 & 0.673 & 0.254 & 21.26 & 0.659 & 0.266 & 20.00 & 0.632 & 0.281 \\
    SUDS \cite{suds} & 22.77 & 0.797 & 0.171 & 23.12 & 0.821 & \textbf{0.135} & 20.76 & 0.747 & 0.198 \\
    MARS \cite{mars} & 24.23 & 0.845 & \textbf{0.160} & 24.00 & 0.801 & 0.164 & 23.23 & 0.756 & \textbf{0.177} \\    
    \midrule
    \multicolumn{1}{c}{Ours} & \textbf{26.59} & \textbf{0.913} & 0.204 & \textbf{26.22} & \textbf{0.907} & 0.207 & \textbf{24.76} & \textbf{0.875} & 0.225 \\
    \bottomrule
  \end{tabular}}
\end{table}

Moreover, the comparative results for the KITTI dataset are depicted in \cref{tab:kitti}. Notably, our method outperforms all other methods based on the PSNR and SSIM metrics, showcasing its effectiveness in capturing high-fidelity reconstructions and preserving structural similarity. However, it is important to note a less favorable performance in terms of the LPIPS metric. The discrepancy in LPIPS results may be attributed to the metric's sensitivity to subtle perceptual differences, which could arise from scene texture and lighting conditions in the KITTI dataset. More results are provided in the supplementary materials.

\subsection{Ablation Studies} 
\label{sec:as}
We validate our method's design decisions through Pandaset experiments, encompassing diverse and challenging autonomous driving scenarios, thoroughly evaluating the impact of the proposed primary components.

\begin{figure}[tb]
  \centering
  \includegraphics[height=3.2cm]{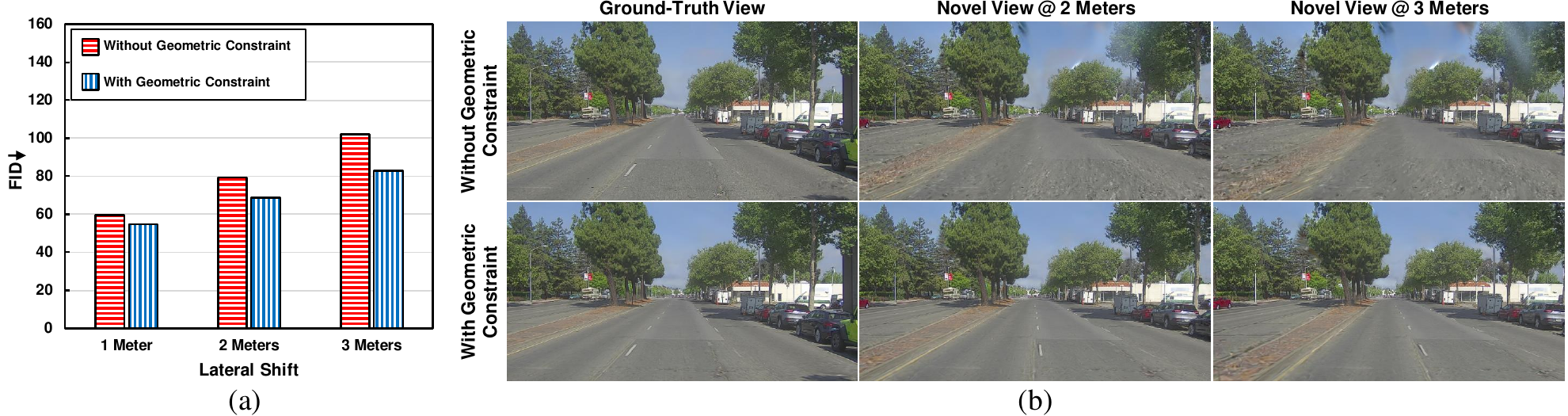}
  \caption{Analysis of the proposed background geometric constraints under various degrees of ego-vehicle lateral shift.}
  \label{fig:bgconstraints}
\end{figure}

\subsubsection{Background Geometry Constraints} 
For this investigation, we tried our method with and without the proposed background geometry constraints on all scenes and measured the average FID based on different amounts of lateral shift. As seen in \cref{fig:bgconstraints} (a), with the proposed geometric constraints, the synthesized sequences exhibit lower FID, indicating better visual quality. Furthermore, the difference between the FID values in these two cases varies from 4.8 to 18.9 points as the amount of lateral shift increases from 1 meter to 3 meters. In \cref{fig:bgconstraints} (b), the qualitative results depicting various degrees of lateral shift are presented. It is evident that, in the absence of geometric constraints, lateral camera displacement leads to significant distortion of the road and the disappearance of lane markings. Additionally, unregulated positioning of the sky Gaussians, while acceptable in ground-truth views, occludes the background during lateral shifts.

\begin{figure}[tb]
  \centering
  \includegraphics[height=3.7cm]{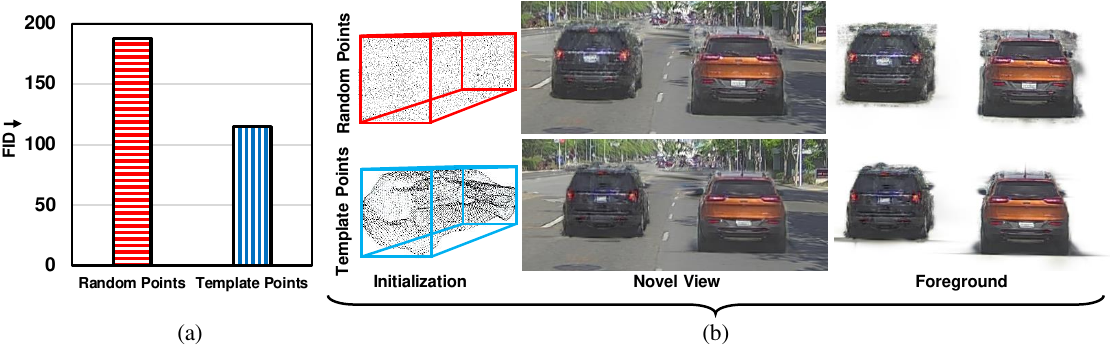}
  \caption{Effect of foreground object Gaussian initialization on novel view synthesis.} 
  \label{fig:box-vs-nfi}
\end{figure}

\subsubsection{Foreground Initialization} 
To validate the effect of using a 3D template for foreground Gaussian initialization, we compared its performance with random initialization in object 3D boxes. Average FID scores of the foreground objects in the novel views obtained by performing a lateral camera shift over all 80 consecutive frames are measured. As depicted in \cref{fig:box-vs-nfi} (a), utilizing template points for initialization results in lower FID scores, indicating that the reconstructed foreground objects bear more similarity to their ground-truth counterparts. This is shown in \cref{fig:box-vs-nfi} (b), in which the foreground objects initialized randomly show box-like distortions around them during novel view synthesis.

\begin{figure}[tb]
  \centering
  \includegraphics[height=1.6cm]{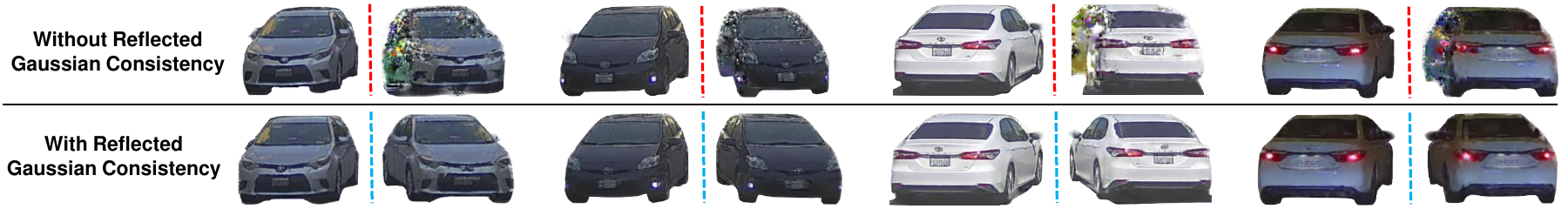}
  \caption{Impact of reflected Gaussian consistency constraint on foreground synthesis.} 
  \label{fig:reflected}
\end{figure}

\subsubsection{Reflected Gaussian Consistency Constraint} We investigated this component, by training our method with and without the reflected Gaussian consistency constraint. Sample results are shown in \cref{fig:reflected}. Under no reflected Gaussian consistency constraint, while the ground-truth views are reconstructed well the symmetrically unseen views contain distortion due to a lack of supervision. On the other hand, the reflected Gaussian consistency constraint harnesses the complete potential of object symmetry properties by mirroring unseen Gaussians along the reflection plane and guiding them with supervision from the ground-truth view. This leads to a more faithful synthesis of the foreground objects from different views.

\begin{figure}[tb]
  \centering
  \includegraphics[height=5.0cm]{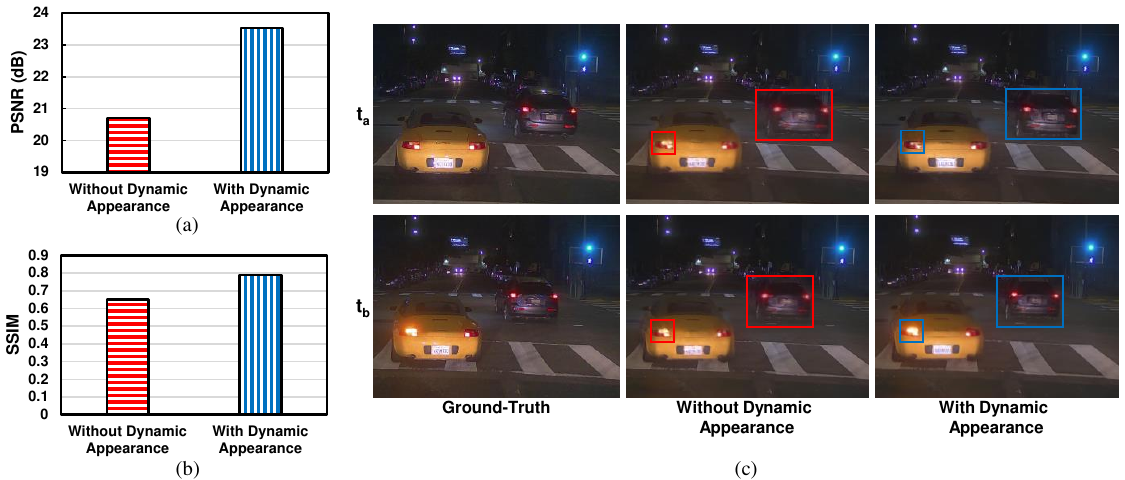}
  \caption{Quantitative and qualitative assessment of dynamic appearance modeling.}
  \label{fig:dynamicmlp}
\end{figure}

\subsubsection{Effect of Dynamic Appearance Modeling} 
We examined the impact of foreground dynamic appearance modeling, as illustrated in \cref{fig:dynamicmlp} (a) and (b). Employing dynamic appearance modeling enhances the quality of reconstructed foreground regions, resulting in higher PSNR and SSIM values. This improvement is attributed to the effective handling of higher frequency details through residual-based modeling, which is not adequately addressed by static appearance modeling. As depicted in \cref{fig:dynamicmlp} (c), this 4D representation not only enhances the quality of foreground objects but also effectively captures the dynamic appearance changes of objects, such as flashing lights.

\begin{figure}[tb]
  \centering
  \includegraphics[height=4.4cm]{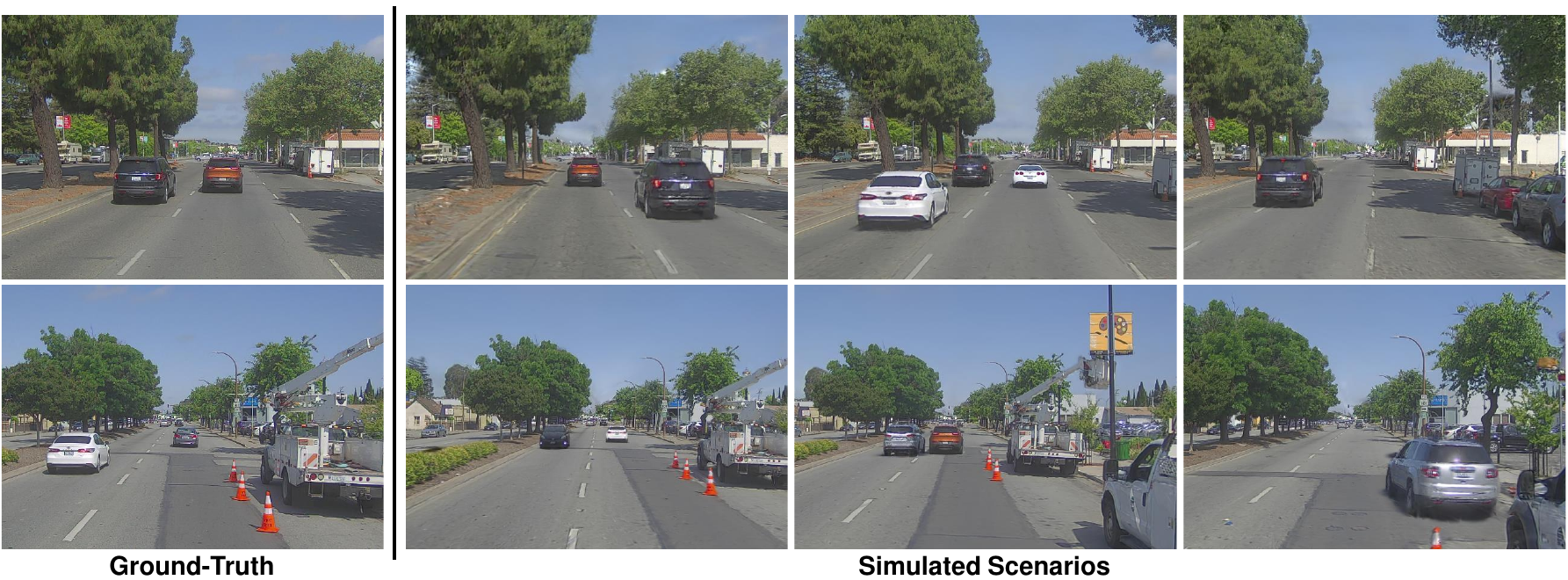}
  \caption{Realistic simulation of diverse autonomous driving scenarios.} 
  \label{fig:simulatedscenario}
\end{figure}

\subsubsection{Novel Scenario Simulation} One of the main benefits of the proposed framework is the ability to simulate novel scenarios. 
In Figure \ref{fig:simulatedscenario}, we present two ground-truth scenes alongside their corresponding simulated scenarios. These scenarios encompass various driving conditions, including ego-vehicle and object lane changes, addition or removal of objects, and simulations of critical events such as a sudden cut-in or collision.

\section{Limitation}
\label{sec:limitation}

Our method is constrained to reconstructing rigid dynamic foreground objects, such as vehicles, and cannot accommodate non-rigid objects including pedestrians, cyclists, \etc. Future work could explore more complex dynamic scene modeling methods to address this limitation. Furthermore, our approach depends on ground-truth 3D boxes and adapts a transformation offset for each object to rectify inaccuracies in their poses. Exploring alternative methods to reduce this dependency, such as leveraging motion information for foreground object identification and trajectory estimation, presents an intriguing avenue for further investigation.

\section{Conclusion}
\label{sec:conclusion}
This paper introduces AutoSplat, a novel approach for accurately reconstructing and synthesizing dynamic autonomous driving scenes. By constraining road and sky Gaussians, we achieve realistic novel view synthesis during ego-vehicle lateral shifts. Exploiting 3D templates for foreground object initialization enables the reflected Gaussian consistency constraint to supervise the symmetrically unseen parts. Additionally, we estimate temporally-dependent, residual spherical harmonics for each foreground object Gaussian to model dynamic appearance. Thorough experiments across various datasets showcases our method's superior performance compared to SOTA methods. Comprehensive ablations demonstrate the effectiveness of our proposed components.

\bibliographystyle{splncs04}
\bibliography{main}

\newpage
\renewcommand{\thesubsection}{\Alph{subsection}}
\section*{Supplementary Materials}
\unnumberedsubsection{Implementation Details}

Our approach builds upon the 3DGS framework \cite{3dgs}, opting for accumulated LiDAR point clouds over conventional SfM points for background initialization. Aligned with the methodology of 3DGS, our background training unfolds across two distinct phases, each spanning 15K iterations and collectively summing up to 30K iterations. Background reconstruction requires masks for the road, sky, and other remaining background regions (excluding foreground objects). We acquire these masks through a pre-trained Mask2Former model \cite{mask2former}. Throughout the two phases of background training, we maintain a fixed positioning of road and sky Gaussians, ensuring their stability and preventing inadvertent reconstruction of other background regions. 

Foreground training spans 5K iterations, during which we utilize masked ground-truth images to supervise the synthesis of foreground objects. To enforce the reflected Gaussian consistency constraint, Gaussians for each foreground object are reflected according to their respective reflection planes every alternate iteration. Subsequently, the rasterized foreground objects with reflected Gaussians undergo supervision using the corresponding masked ground-truth image. 

The foreground training is followed by an additional 10K iterations for scene fusion, wherein both foreground and background Gaussians are fine-tuned together. During fusion, attributes of the foreground object Gaussians are fined-tuned while for background Gaussians the adjustments are solely made to the opacity, as their appearance and geometry are established during the background training phase.

Additionally, our loss terms are configured with $\beta$ and $\gamma$ values set to 1000 and 1, respectively. We increased the grad threshold to 0.001, as a lower threshold significantly increases unnecessary points. All experiments are conducted on a single NVIDIA Tesla V100 GPU with 32 GB of memory.

\begin{figure}[!tb]
  \centering
  \includegraphics[height=13.0cm]{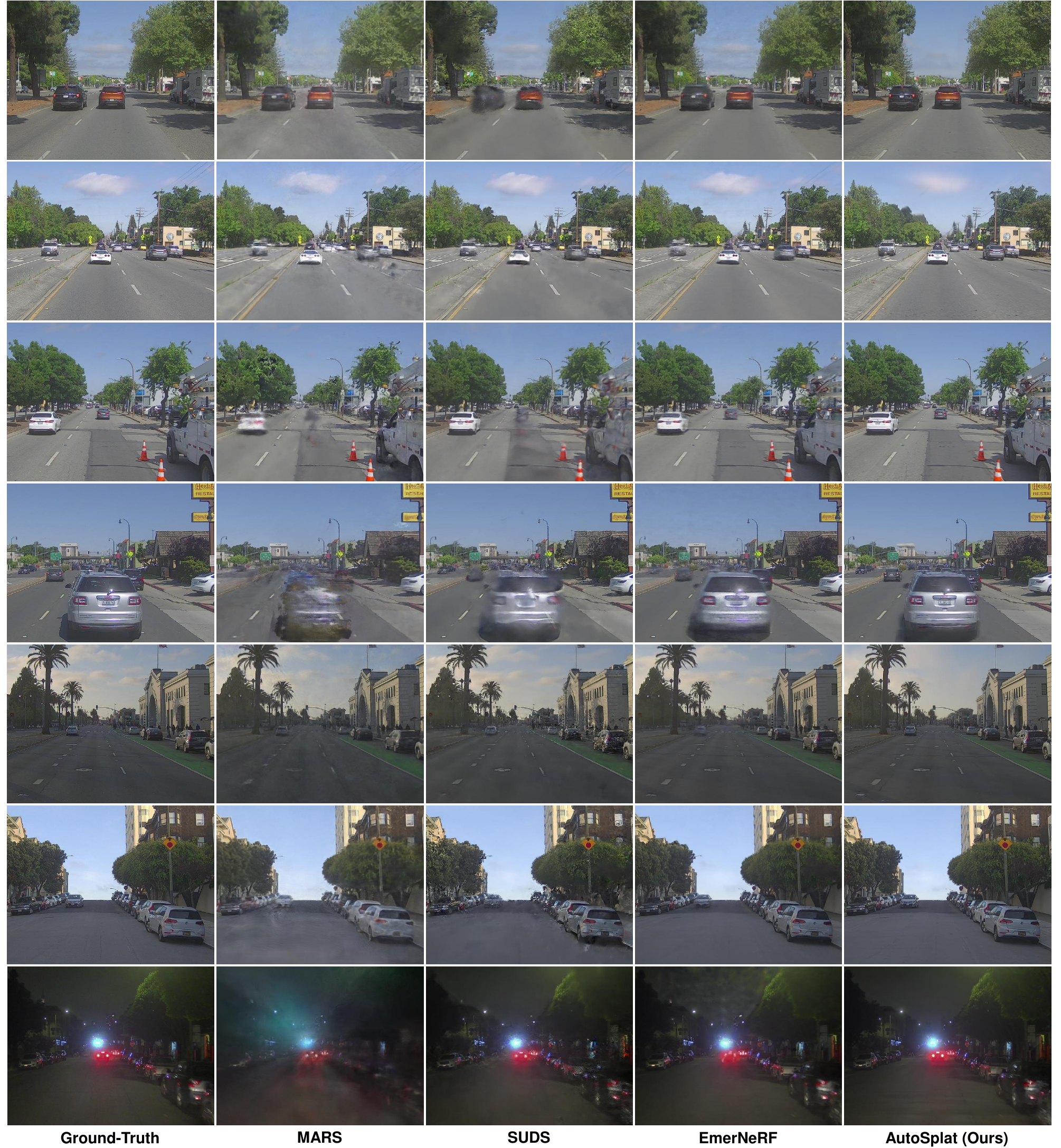}
  \caption{Qualitative comparison of novel view synthesis (test frames) on Pandaset.} 
  \label{fig:pandasetnvs_supp}
\end{figure}

\unnumberedsubsection{More Results}
The qualitative results of the novel view synthesis (test frames) conducted on Pandaset are visually depicted in \cref{fig:pandasetnvs_supp}. The comparative analysis reveals the efficacy of our proposed method, showcasing superior synthesis quality in both background and foreground elements. This demonstrates the robustness and precision of our approach in accurately reproducing complex scenes with intricate details. It is worth noting, while we reconstruct the background with a high degree of realism, our method falls short in accurately modelling details in the sky regions, such as clouds.

\begin{figure}[!tb]
  \centering
  \includegraphics[height=6.3cm]{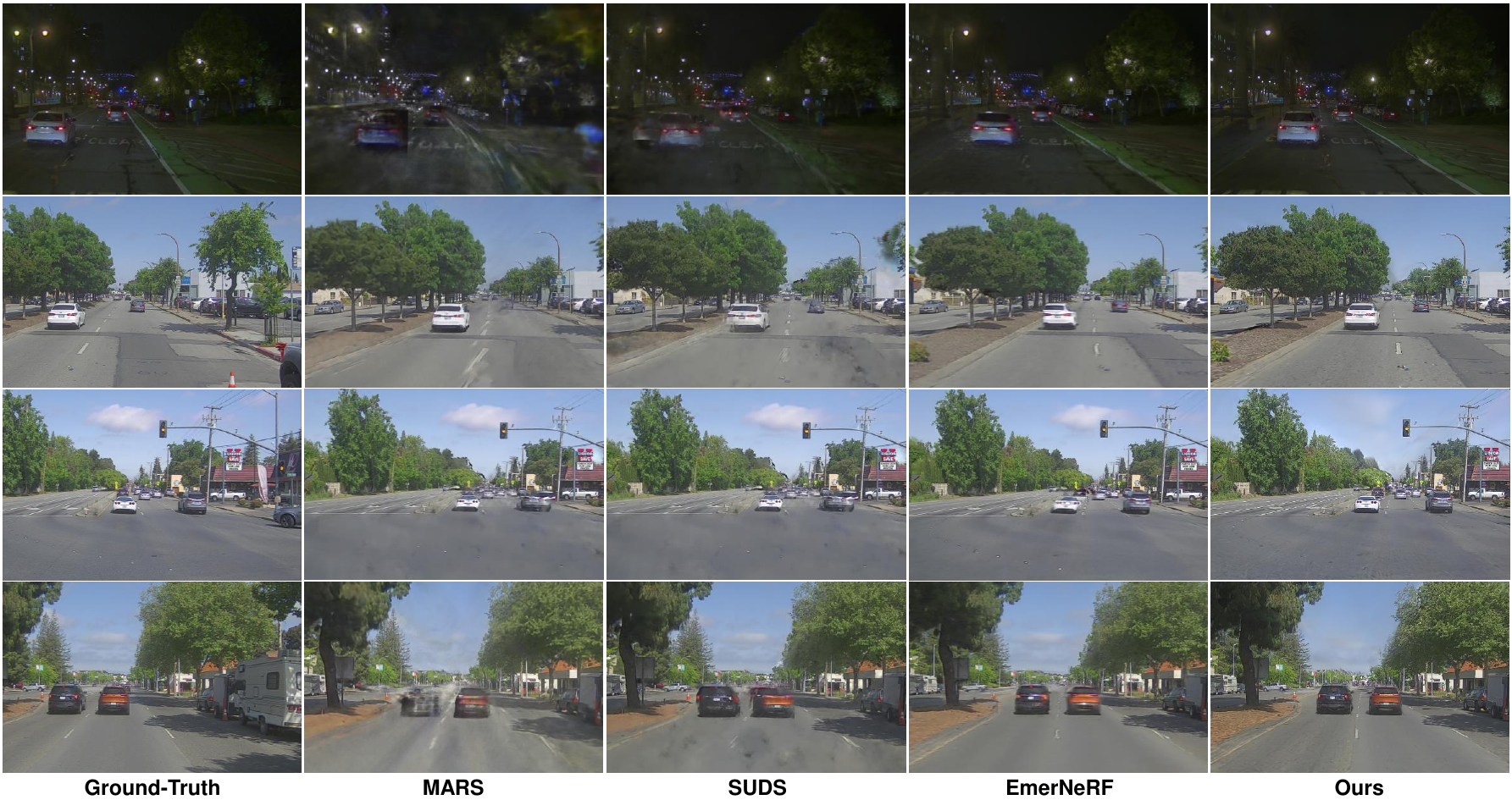}
  \caption{Qualitative comparison of ego-vehicle lateral shift on Pandaset.} 
  \label{fig:pandasetlateralshift_supp}
\end{figure}

In \cref{fig:pandasetlateralshift_supp} more results on ego-vehicle lateral shift are shown. Through the utilization of the background geometry constraint embedded in our approach, we excel in generating backgrounds of exceptional fidelity, preserving intricate elements like road markings. Conversely, competing methods fall short of faithfully reproducing such features, frequently leading to a loss of detail or inconsistent novel view synthesis. Moreover, our method capitalizes on the reflected Gaussian consistency constraint to achieve superior foreground reconstruction, even amidst significant lateral shifts of the ego vehicle. In contrast, alternative techniques struggle to maintain reconstruction quality under similar conditions.

\begin{figure}[!tb]
  \centering
  \includegraphics[height=8.7cm]{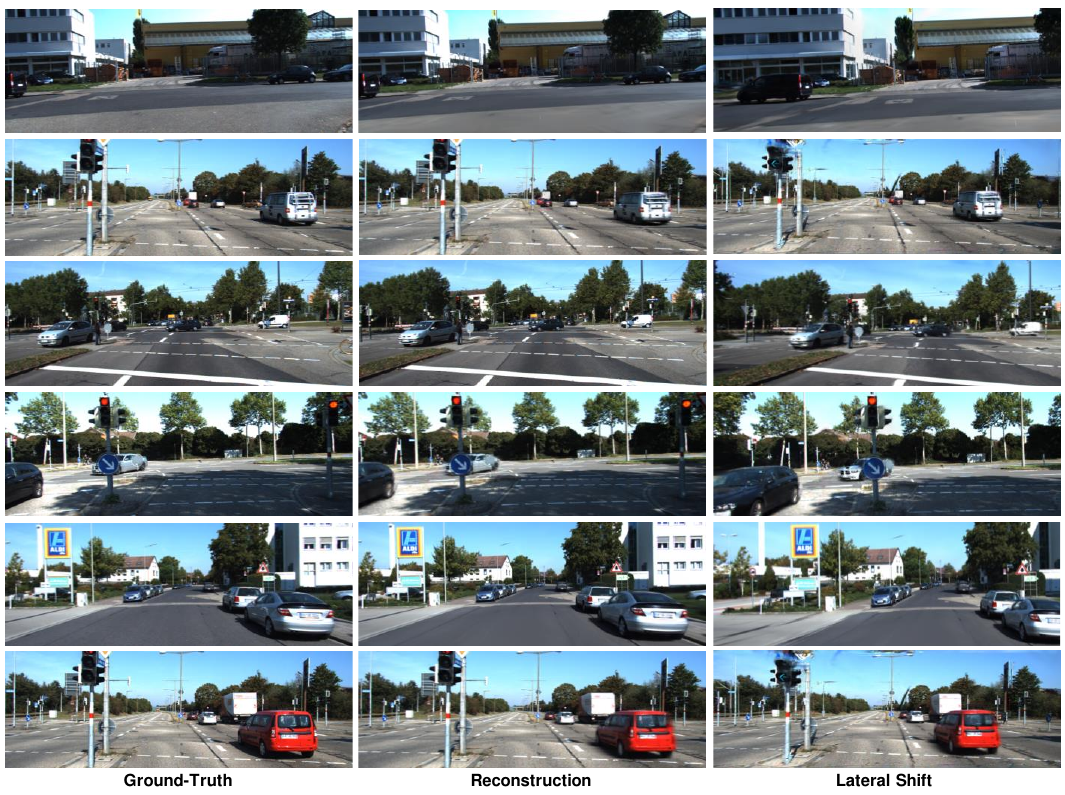}
  \caption{Our scene reconstruction and novel view synthesis results on KITTI.} 
  \label{fig:kitti_recon_supp}
\end{figure}

\begin{figure}[!tb]
  \centering
  \includegraphics[height=7.5cm]{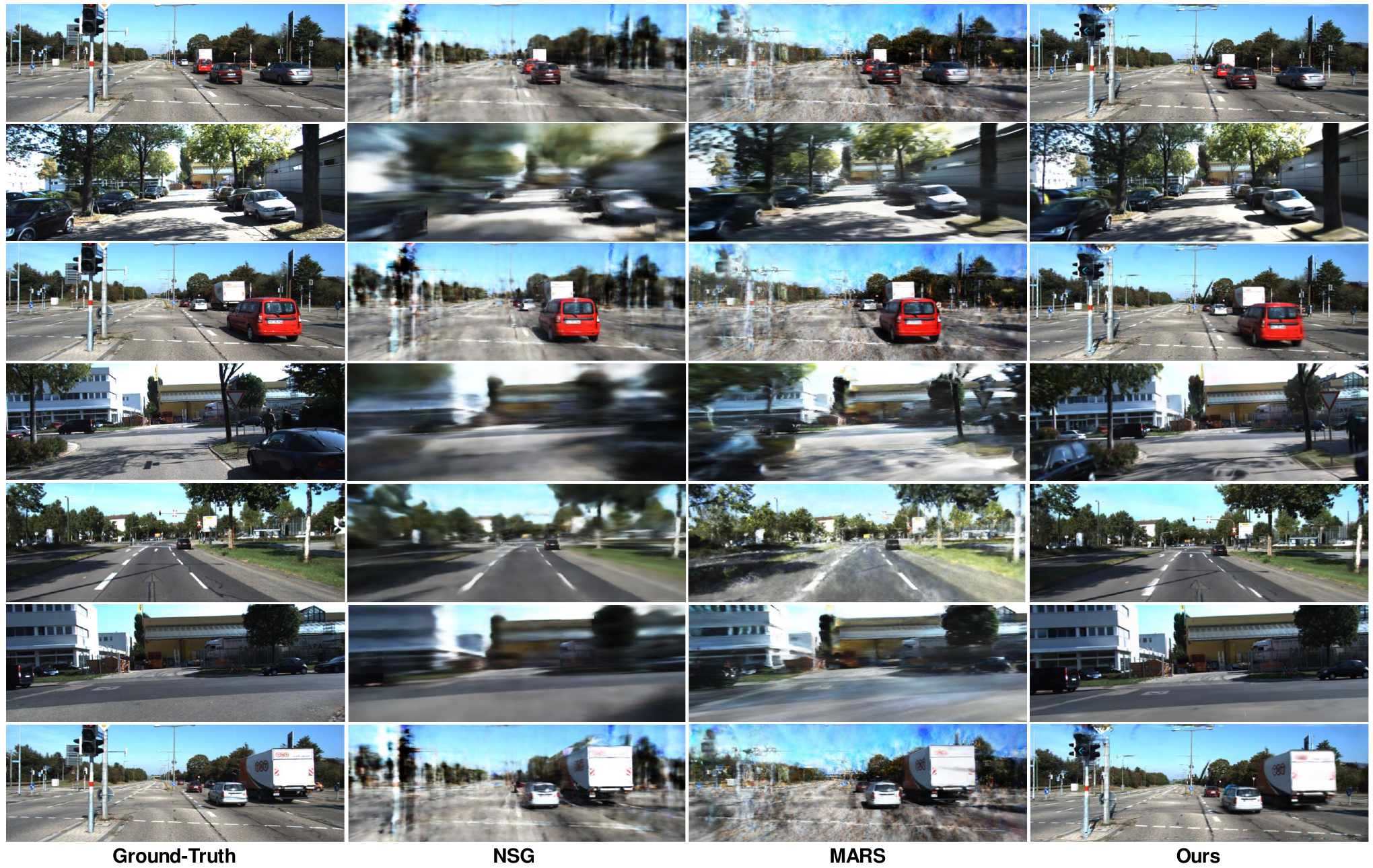}
  \caption{Qualitative comparison of ego-vehicle lateral shift on KITTI.} 
  \label{fig:kitti_lateral_shift_supp}
\end{figure}

In \cref{fig:kitti_recon_supp} qualitative reconstruction and novel view synthesis results on KITTI are shown. As can be seen, our method adeptly reconstructs various autonomous driving scenarios with remarkable fidelity. Additionally, in \cref{fig:kitti_lateral_shift_supp}, a comparison of our method against other methods reveals our superior performance in both background and foreground reconstruction quality when the ego-vehicle undergoes a lateral shift, a common scenario in autonomous driving simulations. Prior approaches produce distortions under novel views while our method's use of geometry constraints during background reconstruction ensures multi-view consistent rasterization.

\begin{figure}[!tb]
  \centering
  \includegraphics[height=3.5cm]{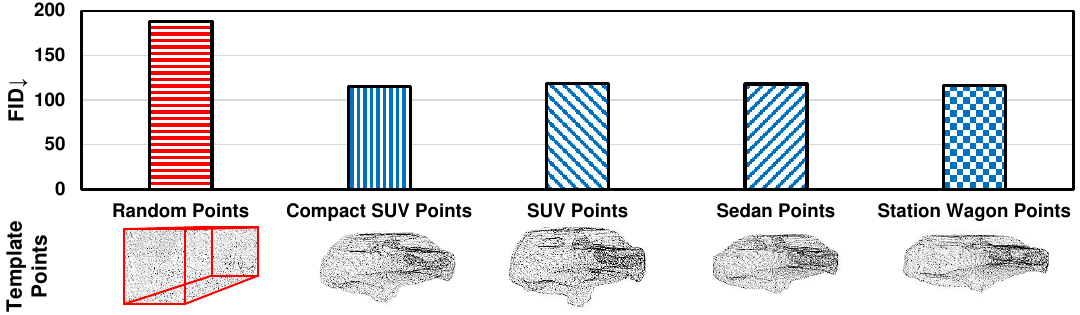}
  \caption{Examining the impact of diverse 3D templates used for foreground object initialization on synthesis quality.} 
  \label{fig:nfi_supp}
\end{figure}

To consider the effect of different 3D templates on reconstruction quality, we performed an experiment in which different types of templates were used for initializing foreground objects. The average FID scores using each 3D template is calculated and shown in \cref{fig:nfi_supp}. The results indicate a consistent decrease in FID values across all templates compared to random initialization, underscoring the efficacy of the proposed initialization method. Notably, while the compact SUV car template yields the lowest FID among all templates, the observed differences are marginal. This is likely due to the fact that the compact SUV is geometrically similar to common cars, having a shape in-between a Sedan and a SUV.
\end{document}